\documentclass{article}

\PassOptionsToPackage{numbers, compress}{natbib}

\usepackage[preprint]{neurips_2025}

\usepackage[utf8]{inputenc} 
\usepackage[T1]{fontenc}    
\usepackage{hyperref}       
\usepackage{url}            
\usepackage{booktabs}       
\usepackage{amsfonts}       
\usepackage{nicefrac}       
\usepackage{microtype}      

\usepackage{caption}
\usepackage{comment}
\usepackage{enumitem}
\usepackage{algorithm2e}
\usepackage{multirow}
\usepackage{csquotes}
\usepackage{graphicx}
\usepackage{subcaption}
\usepackage[table]{xcolor}
\usepackage{amsmath,amssymb}
\usepackage{todonotes}
\usepackage{wrapfig}
\usepackage{algorithmic}
\usepackage{lipsum}
\usepackage[most]{tcolorbox}
\usepackage{xcolor}         
\usepackage{fancyvrb}

\usepackage{amsmath,amsfonts,bm}

\def\eqref#1{equation~\ref{#1}}

\def\1{\bm{1}}

\def\vx{{\bm{x}}}
\def\vy{{\bm{y}}}
\def\vz{{\bm{z}}}

\DeclareMathAlphabet{\mathsfit}{\encodingdefault}{\sfdefault}{m}{sl}
\SetMathAlphabet{\mathsfit}{bold}{\encodingdefault}{\sfdefault}{bx}{n}

\def\gE{{\mathcal{E}}}

\def\sZ{{\mathbb{Z}}}

\newcommand{\Ls}{\mathcal{L}}

\def\passmsg{\textcolor{teal}{\vz_{\mbox{pass}}}}
\def\errormsg{\textcolor{magenta}{\vz_{\mbox{error}}}}

\def\policy{\textcolor{purple}{\bm{\pi}_\theta}}
\def\refpolicy{\textcolor{purple}{\bm{\pi}_{\theta_0}}}
\def\policymix{\ensuremath{\textcolor{purple}{\bm{\pi}}^{\mathrm{sft}}}} 
\def\policymixrl{\ensuremath{\textcolor{purple}{\bm{\pi}}^{\mathrm{rl}}}} 
\def\policymixrldm{\ensuremath{\textcolor{purple}{\bm{\pi}}^{\mathrm{rd}}}} 
\def\policyrag{\ensuremath{\textcolor{cyan}{\bm{\phi}}^{\mathrm{sft}}}} 
\def\policyragrl{\ensuremath{\textcolor{cyan}{\bm{\phi}}^{\mathrm{rl}}}} 
\def\policyragrldm{\ensuremath{\textcolor{cyan}{\bm{\phi}}^{\mathrm{rd}}}} 

\def\FcSftDataset{\ensuremath{\mathbb{D}_{\mathrm{fc}}}}
\def\SpSftDataset{\ensuremath{\mathbb{D}_{\mathrm{sp}}}}
\def\RlDataset{\ensuremath{\mathbb{D}_{\mathrm{rl}}}}

\def\dm{DyMo}
\tcbuselibrary{breakable}

\definecolor{textPreamble}{RGB}{64,84,107}
\definecolor{bgPreamble}{RGB}{184,200,217}

\newtcbox{\preambletitlebox}{
  enhanced,
  colupper=white,
  colback=textPreamble,
  fontupper=\bfseries,
  size=small,
  baseline=3pt,
  nobeforeafter,
  frame code={
    \path[fill=textPreamble] (frame.north west)
    -- ([xshift=2mm]frame.north east)
    -- (frame.south east)
    -- (frame.south west)
    -- (frame.north west)
      [sharp corners]-- cycle;
  }
}

\makeatletter
\newtcolorbox{preamblebox}[1]{
  enhanced,
  breakable,
  skin=bicolor,
  arc=1pt,
  coltitle=white,
  colframe=textPreamble,
  colback=bgPreamble,
  colbacklower=white,
  fonttitle=\bfseries,
  detach title,
  title={#1},
  break at=5mm, 
  pad at break=2mm,
  before upper*={%
    \vskip-\dimexpr\kvtcb@boxsep+\kvtcb@top+.1pt
    \hspace*{-\dimexpr\kvtcb@boxsep+\kvtcb@leftupper+.1pt}%
    \expandafter\preambletitlebox\expandafter{\tcbtitletext} %
    \par\vspace{5pt}\noindent
    \setlength{\parskip}{0.5em}\setlength{\baselineskip}{1.2\baselineskip}\setlength{\parindent}{0pt}
  }
}
\makeatother

\definecolor{textPrompt}{RGB}{124, 104, 173}
\definecolor{bgPrompt}{RGB}{215,206,228}

\newtcbox{\prompttitlebox}{
  enhanced,
  colupper=white,
  colback=textPrompt,
  fontupper=\bfseries,
  size=small,
  baseline=3pt,
  nobeforeafter,
  frame code={
    \path[fill=textPrompt] (frame.north west)
    -- ([xshift=2mm]frame.north east)
    -- (frame.south east)
    -- (frame.south west)
    -- (frame.north west)
      [sharp corners]-- cycle;
  }
}

\makeatletter
\newtcolorbox{promptbox}[1]{
  enhanced,
  breakable,
  skin=bicolor,
  arc=1pt,
  coltitle=white,
  colframe=textPrompt,
  colback=bgPrompt,
  colbacklower=white,
  fonttitle=\bfseries,
  detach title,
break at=5mm, 
  pad at break=2mm,
  title={#1},
  before upper*={%
    \vskip-\dimexpr\kvtcb@boxsep+\kvtcb@top+.1pt
    \hspace*{-\dimexpr\kvtcb@boxsep+\kvtcb@leftupper+.1pt}%
    \expandafter\prompttitlebox\expandafter{\tcbtitletext} %
    \par\vspace{5pt}\noindent
    \setlength{\parskip}{0.5em}\setlength{\baselineskip}{1.2\baselineskip}\setlength{\parindent}{0pt}
  }
}
\makeatother

\definecolor{textCompletion}{RGB}{181,142,84}
\definecolor{bgCompletion}{RGB}{236,218,172}

\newtcbox{\completiontitlebox}{
  enhanced,
  colupper=white,
  colback=textCompletion,
  fontupper=\bfseries,
  size=small,
  baseline=3pt,
  nobeforeafter,
  frame code={
    \path[fill=textCompletion] (frame.north west)
    -- ([xshift=2mm]frame.north east)
    -- (frame.south east)
    -- (frame.south west)
    -- (frame.north west)
      [sharp corners]-- cycle;
  }
}

\makeatletter
\newtcolorbox{completionbox}[1]{
  enhanced,
  breakable,
  skin=bicolor,
  arc=1pt,
  coltitle=white,
  colframe=textCompletion,
  colback=bgCompletion,
  colbacklower=white,
  fonttitle=\bfseries,
  detach title,
  break at=5mm, 
  pad at break=2mm,
  title={#1},
  before upper*={%
    \vskip-\dimexpr\kvtcb@boxsep+\kvtcb@top+.1pt
    \hspace*{-\dimexpr\kvtcb@boxsep+\kvtcb@leftupper+.1pt}%
    \expandafter\completiontitlebox\expandafter{\tcbtitletext} %
    \par\vspace{5pt}\noindent
    \setlength{\parskip}{0.5em}\setlength{\baselineskip}{1.2\baselineskip}\setlength{\parindent}{0pt}
  }
}
\makeatother

\definecolor{textErrorResult}{RGB}{216,160,199}
\definecolor{bgErrorResult}{RGB}{247,238,246}

\newtcbox{\errortitlebox}{
  enhanced,
  colupper=white,
  colback=textErrorResult,
  fontupper=\bfseries,
  size=small,
  baseline=3pt,
  nobeforeafter,
  frame code={
    \path[fill=textErrorResult] (frame.north west)
    -- ([xshift=2mm]frame.north east)
    -- (frame.south east)
    -- (frame.south west)
    -- (frame.north west)
      [sharp corners]-- cycle;
  }
}

\makeatletter
\newtcolorbox{errorbox}[1]{
  enhanced,
  breakable,
  skin=bicolor,
  arc=1pt,
  coltitle=white,
  colframe=textErrorResult,
  colback=bgErrorResult,
  colbacklower=white,
  fonttitle=\bfseries,
  detach title,
  break at=5mm, 
  pad at break=2mm,
  title={#1},
  before upper*={%
    \vskip-\dimexpr\kvtcb@boxsep+\kvtcb@top+.1pt
    \hspace*{-\dimexpr\kvtcb@boxsep+\kvtcb@leftupper+.1pt}%
    \expandafter\errortitlebox\expandafter{\tcbtitletext} %
    \par\vspace{5pt}\noindent
    \setlength{\parskip}{0.5em}\setlength{\baselineskip}{1.2\baselineskip}\setlength{\parindent}{0pt}
  }
}
\makeatother

\definecolor{textPassResult}{RGB}{100,130,100}
\definecolor{bgPassResult}{RGB}{220,233,218}

\newtcbox{\passtitlebox}{
  enhanced,
  colupper=white,
  colback=textPassResult,
  fontupper=\bfseries,
  size=small,
  baseline=3pt,
  nobeforeafter,
  frame code={
    \path[fill=textPassResult] (frame.north west)
    -- ([xshift=2mm]frame.north east)
    -- (frame.south east)
    -- (frame.south west)
    -- (frame.north west)
      [sharp corners]-- cycle;
  }
}

\makeatletter
\newtcolorbox{passbox}[1]{
  enhanced,
  breakable,
  skin=bicolor,
  arc=1pt,
  coltitle=white,
  colframe=textPassResult,
  colback=bgPassResult,
  colbacklower=white,
  fonttitle=\bfseries,
  detach title,
  break at=5mm, 
  pad at break=2mm,
  title={#1},
  before upper*={%
    \vskip-\dimexpr\kvtcb@boxsep+\kvtcb@top+.1pt
    \hspace*{-\dimexpr\kvtcb@boxsep+\kvtcb@leftupper+.1pt}%
    \expandafter\passtitlebox\expandafter{\tcbtitletext} %
    \par\vspace{5pt}\noindent
    \setlength{\parskip}{0.5em}\setlength{\baselineskip}{1.2\baselineskip}\setlength{\parindent}{0pt}
  }
}
\makeatother

\SetKwComment{Comment}{/* }{ */}
\RestyleAlgo{ruled}
\SetKwInOut{KwArg}{Given}

\title{World Modelling Improves Language Model Agents} %

\author{%
  Shangmin Guo\thanks{Correspondence author: \texttt{s.guo@ed.ac.uk}}\hspace{.5em}\thanks{Work done during internship at Cohere.} \\
  University of Edinburgh\\
  \And
  Omar Darwiche Domingues \\
  Cohere \\
  \And
  Rapha\"{e}l Avalos\footnotemark[2]\\
  Vrije Universiteit Brussel\\
  \And
  Aaron Courville\\
  Université de Montréal\\
  \And
  Florian Strub\\
  Cohere\\
}

\begin{document}

\maketitle

\begin{abstract}
  Tool use in stateful environments presents unique challenges for large language models (LLMs), where existing test-time compute strategies relying on repeated trials in the environment are impractical.
We propose dynamics modelling (\dm), a method that augments LLMs with a state prediction capability alongside function calling during post-training.
This enables LLMs to predict the future states of their actions through an internal environment model.
On the Berkeley Function Calling Leaderboard V2, \dm~improves success rates and significantly reduces hallucinations.
We further integrate the internal environment model into self-verification sampling (SVS), and show that this substantially improves pass\textasciicircum$k$ over number of trials $k$, and allows the model to refuse unreliable outputs.
Together, \dm~and SVS greatly enhance the effectiveness and reliability of LLMs for tool use.
We believe this work charts a path towards scalable planning RL methods for LLM inference without repeatedly querying the oracle environment.

\end{abstract}

\section{Introduction}
\label{sec:introduction}

Large language models (LLMs) have demonstrated remarkable performance in a wide range of applications~\citep{achiam2023gpt4,team2023gemini, meta2024llama, yang2024qwen2, liu2024deepseek, cohere2025command}.
In addition to conventional natural language tasks, recent advances have shown that LLMs also achieve breakthrough performance in formal language tasks, notably code generation~\citep{jimenez2024swebench, gehring2024rlef, claude3.7} and tool use~\citep{cohere2024tooluse, yao2023react, gpt2023tooluse}.
Recent work has shown that scaling the test-time compute can further improve the performance of LLMs on complex tasks such as mathematical reasoning~\citep{shinn2023reflexion,lightman2023let,villalobos2024trading,snell2025scaling,xiong2025self}.
To achieve better performance by scaling up test-time compute, existing methods assume that a verifier, e.g. a process reward model (PRM) or an outcome reward model (ORM), can be queried multiple times during inference~\cite{yao2023react, shinn2023reflexion, snell2025scaling, lightman2023let}.

However, many real-world applications may not rely on a verifier to improve test-time sampling, especially when the LLM interacts with the world as in Agentic scenarios.
One may not execute $k$ payments and be satisfied that one of the payments is correct among the $k$ ones, whereas one may verify $k$ times if a mathematical solution is correct without ramifications.
In this spirit, we consider tool-use tasks, where the agent must execute a single trajectory.
In particular, such constraints are often inherent to the \textbf {statefulness} of environments, i.e., the environment status states after executing an action and cannot be easily reverted - the bank account is reduced after a payment! 

\begin{figure}[!ht]
    \centering
    \includegraphics[width=\textwidth]{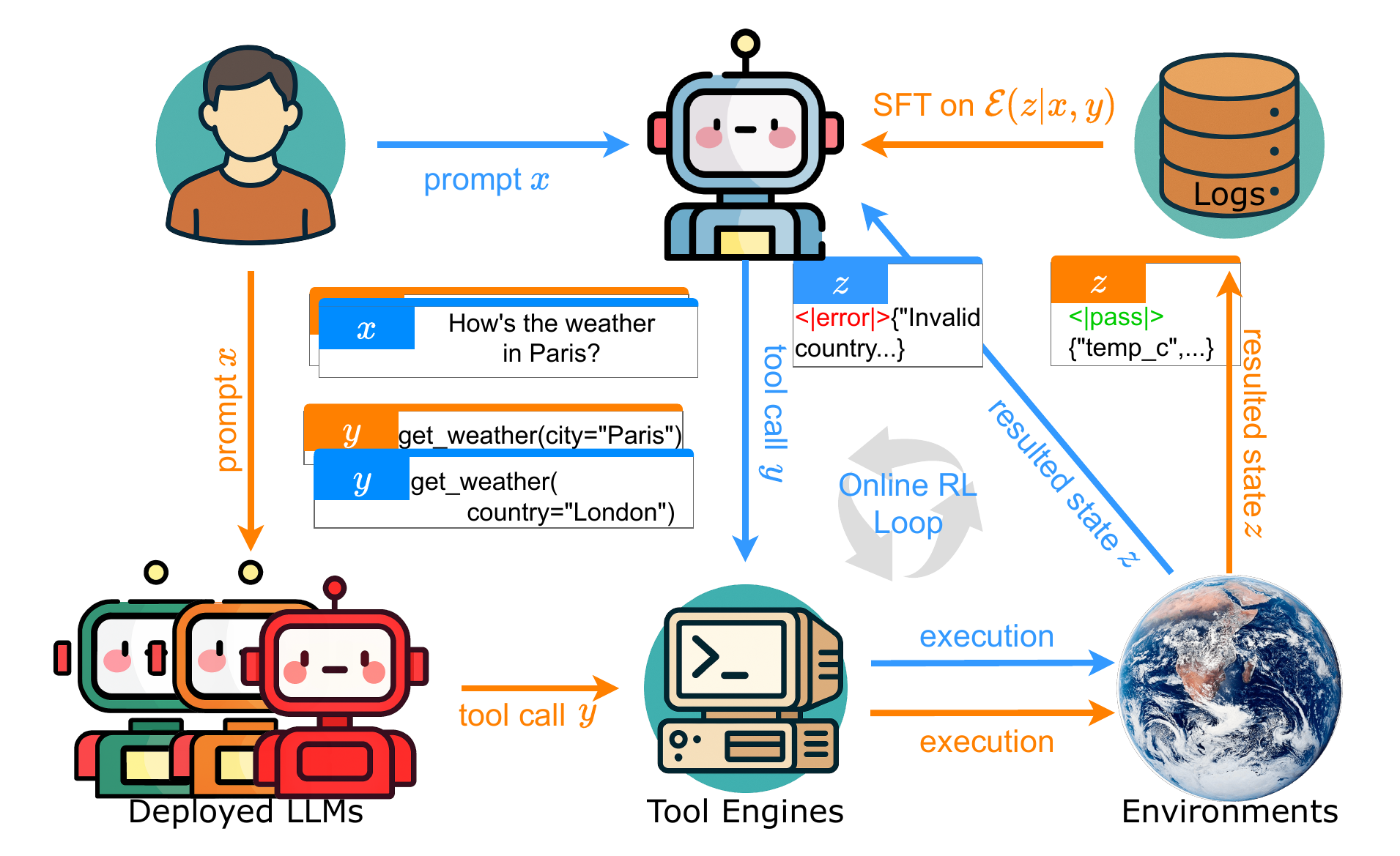}
    \caption{The proposed dynamics modelling (\dm) that trains an LLM to predict the resulted states of the environment after tools execute function calls via either \textcolor{orange}{SFT (orange arrows) on run-logs} or \textcolor{blue}{over online RL loops (blue arrows)}. 
    }
    \label{fig:intro:dymo}
\end{figure}


Inspired by the Generative Verifier~\cite{zhang2024generative} (GenRM), which formulates the reward function as a next token prediction task, as illustrated in Figure~\ref{fig:intro:dymo}, we propose \textbf{dynamic modelling (\dm)} to fine-tune LLMs to generate not only the functions calls for a given user prompt, but also the \emph{subsequent states} of tool engines after executing the generated function calls.
This has two advantages: 1) at training time, this state prediction provides an additional training signal; 2) at test time, this state prediction can be used in the decision-making process to execute the roll-out, similar to one-step planning methods.



We first explore the impact of \dm~into both the supervised fine-tuning (SFT) and RL stages in LLM post-training ~\cite{meta2024llama, yang2024qwen2, liu2024deepseek}, and investigate its effectiveness.
Our results on the Berkeley Function Calling Leaderboard~\cite{bfcl2024} V2 (BFCL-V2) show that \dm~ alleviates the hallucination problem of the SFTed model, and improves the success rate of the RLed models.
Incorporating \dm, our results suggest that an 8B model, when given access to the environment during training, can match and occasionally surpass the performance of GPT-4o on BFCL-V2.

Second, we explore the planning capabilities of \dm~through \textbf{self-verification sampling (SVS)} strategy~\cite{zhao2025sample} at test time.
Specifically, the models generate $k$ tool calls for a given user prompt, predict the respective states resulting from those actions, and proceed with the most promising trajectory based on a ranking mechanism: \emph{sample}, \emph{predict}, then \emph{proceed}.
%

\begin{figure}[!ht]
    \centering
    \includegraphics[width=\textwidth]{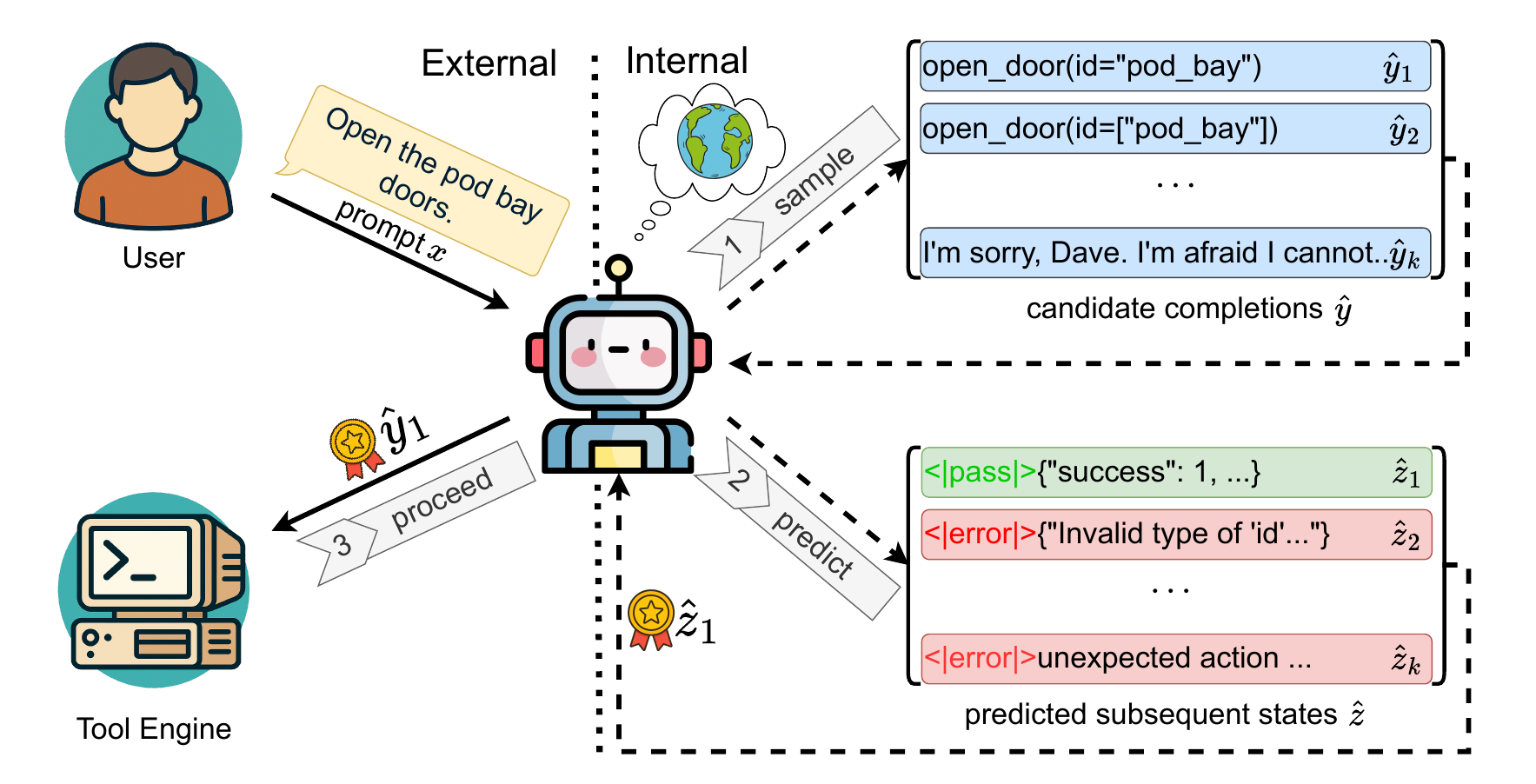} 
    \caption{
        The proposed self-verification sampling (SVS) strategy for test-time compute scaling.
        For a given user prompt $\vx$, the model:
        1) generates $k$ candidate completions $\hat{\vy}$;
        2) predicts the subsequent states $\hat{\vz}$ of the $k$ candidates;
        3) selects a completion to output by a specified scoring function $\textit{score}$.
    }
    \label{fig:intro:svs}
\end{figure}

Our experiments demonstrate that (i) increasing the number of trajectories keeps increasing the LLMs score, (ii) the outcome of the state prediction can be used to select a successful trajectory \emph{without} access to the oracle environment, thereby offering a novel schema for scaling test-time compute in stateful environments.
Furthermore, SVS enables models to effectively \emph{``refuse''} requests that exceed their capabilities based on their state prediction, substantially improving the precision of the final outputs.
We interpret this precision as \emph{``reliability''}, as it represents the proportion of outputs verified as correct by the oracle environment.

In summary, the proposed \dm~method coupled with the SVS strategy significantly enhances the success rate and reliability of LLMs in tool use tasks.

\newpage
\section{Background}
\label{sec:background}

\vspace{-0.5em}
\textbf{Tool Use by LLMs}:
Recent works have demonstrated the capability of LLMs to achieve notable performance in API usage through supervised fine-tuning (SFT) using demonstrations provided either by human experts or generated by advanced models such as GPT-4~\cite{li2023apibank, qin2024toolllm, liu2025toolace}.
This capability positions LLMs as back-ends for agents interacting with environments consisting of various tools~\cite{bfcl2024, liu2024agentbench, apple2024toolsandbox} and simulated user interactions~\cite{yao2025taubench}.
However, existing approaches are mainly based on imitation learning for training~\cite{li2023apibank, qin2024toolllm, liu2025toolace}, while the evaluation relies on interactions between environments and LLMs~\cite{yao2025taubench, bfcl2024}.
Similar to some recent works~\cite{yu2024steptool,chen2025rlagent}, we focus on learning directly through interacting with the environments, as detailed in Section~\ref{ssec:method:dm}.



\textbf{Reinforcement Learning for Fine-tuning LLMs}:
Existing RL methodologies for fine-tuning LLMs primarily address alignment tasks~\cite{stiennon2020rlhf, ouyang2022training, dong2024rlhf} or reasoning-oriented tasks, such as mathematics and programming challenges~\cite{guo2025deepseek}.
Nonetheless, we posit that RL techniques can effectively extend to tool use scenarios, especially when scaling the quantity of generated tool interactions, given that LLMs have already achieved promising performance in real-world tool use tasks~\cite{bfcl, apple2024toolsandbox, yao2025taubench}.
Furthermore, recent studies indicate a substantial performance gap between online/on-policy RL methods and their offline/off-policy counterparts~\cite{xiong2023iterative, guo2024oaif, tang2024understanding, noukhovitch2025faster}.
Although rigorous online interactions can be traded for enhancing wall-clock efficiency, strictly online RL methods still represent an optimal Pareto frontier~\cite{noukhovitch2025faster, bartoldson2025trajectory}.
Hence, to fully harness the capabilities of RL in tool use contexts, our experiment setup is strictly online and on-policy in this work.
Additionally, our method enables models to do one-step planning based their internal learnt environment model during inference time, as illustrated in Section~\ref{ssec:method:svs}.



\textbf{Test-time Compute Scaling}:
It is well-established that LLMs enhance their performance on logical reasoning tasks by generating extended responses that include explicit intermediate reasoning steps~\cite{wei2022chain}.
Further research highlights the importance of explicitly learning these intermediate reasoning stages guided by Policy Reward Models (PRMs) to achieve superior outcomes~\cite{lightman2023let}. 
While scaling test-time computes by lengthening generated completions has proven beneficial~\cite{jaech2024openai, snell2025scaling}, environmental interactions remain critical for achieving optimal results in agent-based tasks~\cite{yao2023react}.
Recent advances also investigate multiple self-rewarding~\cite{xiong2025self}, or self-verification~\cite{zhao2025sample} steps to scale test-time compute in mathematical reasoning contexts.
Unlike these works which query the environment multiple times during inference, we propose to utilise the internal environment models of LLMs to increase the number of completions for scaling test-time computes, as introduced in Section~\ref{ssec:method:svs}.

\section{Methodology}
\label{sec:method}

\subsection{Formulation}
\label{ssec:method:formulation}

We used pre-trained Transformer~\cite{vaswani2017attention} models $\policy$ parameterised by $\theta$ that predict tokens in an autoregressive manner.
After post-training by SFT and RL and given a user prompt $\vx$, the models can then generate completions/responses $\vy$ from the distribution $\vy \sim \policy(\cdot|\vx)$.
Since we focus on the tool use scenario in this work, we assume the user prompts $\vx$ are all about requesting function calls, whereas the completions $\vy$ can be either natural languages or formatted formal languages.
For completions that call functions, they will then be passed to the environment $\gE$ to execute, and the resulted state is $\vz = \gE(\vx, \vy)$ whose complete set is $\sZ$.
Note that using no-tool environment is sometimes available in some experiments measuring hallucination.

Following RL terminology, we refer to $\vx$ as the input \emph{state}, $\vy$ as the generated \emph{action} from the model $\policy$, and $\vz$ as the resulted next state.
The transition dynamics are specified by the environment $\gE$, and the reward function $r:\sZ\mapsto[0, 1]$ assigns a binary score to a pair $(\vx, \vy)$ according to their resultant state $\vz$.
From this RL perspective, our model $\policy$ can:
\begin{itemize}
    \item generate a tool call (action) $\vy$ given a user prompt (state) $\vx$ as input state, i.e. $\vy \sim \policy(\cdot|\vx)$;
    \item predict next-state $\vz$ given a user prompt $\vx$ and a tool call $\vy$, i.e. $\vz\sim\policy(\cdot|\vx,\vy)$.
\end{itemize}



\subsection{\dm: Dynamics Modelling}
\label{ssec:method:dm}

The learning objective of the proposed \dm~is not only the tool use function but the environment function $\gE$.
As illustrated below, we introduce the \dm~into both the SFT and RL stages.

\subsubsection{Dynamics Modelling by Supervised Fine-tuning}
\label{sssec:method:dm:sft}

During the SFT stage, we construct two distinct datasets — one for the tool use function and one for the environment function — which are described in detail below.

For the tool use function, we train the model $\policy$ on a dataset of function calls represented by function call (fc) pairs in the form \verb+<prompt,completion>+, i.e. $\FcSftDataset = \left\{ (\vx_i, \vy_i) \right\}_{i=1}^{N_{\mathrm{fc}}}$.
To train the model on these pairs, we minimise the cross-entropy loss~\cite{goodfellow2016deep} of the model's completion prediction distribution $\policy(\cdot|\vx)$ over them:
\begin{equation}
    \Ls_{\text{FC}} \left(\FcSftDataset;\textcolor{blue}{\theta}\right) = - \sum_{i=1}^{N_{\mathrm{fc}}} \sum_{t=1}^{T_{\vy_i}} \log \policy\bigl(y_{i,t} | \vx_i, \vy_{i,<t}\bigr)
    \label{eq:method:dm:sft:fc_loss}
\end{equation}
where $y_{i,t}$ is the $t$-the element in the target completion $\vy_i$, $\vy_{i,<t}$ represents the partial target sequence preceding $y_t$, and $T_{\vy_i}$ is the length of $\vy_i$.

Regarding the environment function, we represent it by a dataset of state prediction (sp) triplets in the form \verb+<prompt,completion,result>+, i.e. $\SpSftDataset = \left\{ (\vx_i, \vy_i, \vz_i) \right\}_{i=1}^{N_{\mathrm{sp}}}$.
Such data can be gathered and curated from the accumulated \emph{run logs} of the target environment function $\gE$, which we argue is a under-explored source for data scaling.
Similar to the tool use function, we minimise the cross-entropy loss of the model's state prediction distribution $\policy(\cdot|\vx,\vy)$ over these triplets:
\begin{equation}
    \Ls_{\text{SP}}(\SpSftDataset;\textcolor{blue}{\theta}) = -\sum_{i=1}^{N_{\mathrm{sp}}} \sum_{t=1}^{T_{\vz_i}} \log \policy\bigl(z_{i,t} | \vx_i, \vy_i, \vz_{i,<t}\bigr)
    \label{eq:method:dm:sft:sp_loss}
\end{equation}
where the indices $i$ and $t$ follow the same meanings as in Equation~\ref{eq:method:dm:sft:fc_loss}, and $T_{\vz_i}$ is the length of $\vz_i$.

\subsubsection{Dynamics Modelling over Online Reinforcement Learning}
\label{sssec:method:dm:rl}

In addition to the SFT stage, \dm~can also be incorporated into the RL fine-tuning of LLMs.

Starting from a prompt set $\RlDataset=\{\vx_i\}_{i=1}^{N_{\mathrm{rl}}}$, we first sample \emph{two} completions from the model, i.e. $\hat{\vy}_i^1,\hat{\vy}_i^2\sim\policy(\cdot|\vx_i)$.
The two completions along with the prompt $\vx_i$ are then passed as inputs to the environment function to get the next states, i.e. $\vz_i^1=\gE(\vx_i, \hat{\vy}_i^1)$ and $\vz_i^2=\gE(\vx_i, \hat{\vy}_i^2)$.
Binary scores are then assigned to the \verb+<prompt,completion>+ pairs by the reward function $r$, i.e. $r_1 = r(\vx_i, \hat{\vy}_i^1)$ and $r_2 = r(\vx_i, \hat{\vy}_i^2)$.
Subsequently, we sample predicted next states $\hat{\vz}_i^1$ and $\hat{\vz}_i^2$ from the model, i.e. $\hat{\vz}_i^1\sim\policy(\cdot|\vx_i,\hat{\vy}_i^1)$ and $\hat{\vz}_i^2\sim\policy(\cdot|\vx_i,\hat{\vy}_i^2)$, to track the state prediction performance.
Per RL training step, we update the parameter $\theta$ of the model $\policy$ to simultaneously minimise the online two-sample REINFORCE Leave-One-Out (RLOO) loss~\cite{ahmadian2024back,flet2024copg} given in Equation~\ref{eq:method:dm:rl:copg_loss}, and the cross-entropy sample loss given in Equation~\ref{eq:method:dm:rl:sft_loss}.

\begin{equation}
    \begin{aligned}
        \Ls_{\text{RLOO}}(\RlDataset; \textcolor{blue}{\theta}) = -\sum_{i=1}^{N_{\mathrm{rl}}} \biggl[ 
        & \left( r_{\beta/2}^{\policy}(\vx_i,\hat{\vy}_i^1) - r_{\beta/2}^{\policy}(\vx_i,\hat{\vy}_i^2) \right)
          \log \left( \policy(\hat{\vy}_i^1|\vx_i) \right) \\
        + & \left( r_{\beta/2}^{\policy}(\vx_i,\hat{\vy}_i^2) - r_{\beta/2}^{\policy}(\vx_i,\hat{\vy}_i^1) \right)
          \log \left( \policy(\hat{\vy}_i^2|\vx_i) \right) \biggr]
    \end{aligned}
    \label{eq:method:dm:rl:copg_loss}
\end{equation}

\begin{equation}
    \Ls_{\text{DM}}(\RlDataset; \textcolor{blue}{\theta}) =
    -\sum_{i=1}^{N_{\mathrm{rl}}} \sum_{j=1}^{2} \sum_{t=1}^{T_{\hat{\vz}_i^j}}
    \log \policy \bigl(z_{i,t}^j | \vx_i, \hat{\vy}_i^j, \vz_{i,<t}^j \bigr)
    \label{eq:method:dm:rl:sft_loss}
\end{equation}

where $\theta_0$ is the detached initial parameter in the RL stage, $\beta$ is a constant hyperparameter, and $r_{\beta/2}^{\policy}(\vy_i)$ is the regularised reward defined as $r_j-\frac{\beta}{2}\log\frac{\policy(\vy_i^j|\vx_i)}{\refpolicy(\vy_i^j|\vx_i)}$ for $j\in\{1,2\}$.



\subsection{Self-Verification Sampling by Internal Environment Model}
\label{ssec:method:svs}

\begin{wrapfigure}{R}{0.6\textwidth}
    \vspace{-1em}
    \begin{minipage}{0.6\textwidth}
        \begin{algorithm}[H]
            \caption{Self-verification sampling (SVS)}
            \KwIn{$\vx$, number of candidate completions $k$}
            \KwOut{a completion $\hat{\vy}$}
            \KwArg{a pre-specified scoring function $\textit{score}$}
            \For{$i\gets1$ \KwTo $k$}{
                $\hat{\vy}_i\sim\policy(\cdot|\vx)$\;
                $\hat{\vz}_i\sim\policy(\cdot|\vx,\hat{\vy}_i)$\;
            }
            $j\gets \textit{score} \bigl(\policy(\hat{\vz}_1|\vx_1, \hat{\vy}_1),\dots,\policy(\hat{\vz}_k|\vx_k, \hat{\vy}_k)\bigr)$\;
            $\hat{\vy}\gets\hat{\vy}_j$\;
            \label{alg:method:svs}
        \end{algorithm}
    \end{minipage}
\end{wrapfigure}

After undergoing \dm~in both the SFT and RL phases, our model $\policy$ is capable of both generating tool calls and predicting the subsequent states after executing them.
Leveraging this capability, we propose to query the internal environment model of $\policy$ multiple times to do Self-Verification Sampling (SVS), as illustrated in Algorithm~\ref{alg:method:svs}.
Given multiple completions per prompt, SVS selects a single output based on a specified scoring function $\textit{score}$ and the internal environment model of $\policy$.
Notably, unlike existing approaches, SVS scales test-time compute \emph{without} querying the oracle environment $\gE$.
This approach is reminiscent of the Best-of-$N$ search strategy described in~\cite{snell2025scaling}, but avoids querying the environment $\gE$ multiple times, thereby preventing unintended state changes caused by repeated trials.
In addition, SVS aligns with the notion of mental simulation in decision-making, a concept explored in cognitive science~\cite{klein2018role}, thereby establishing a conceptual bridge between research in RL and cognitive science\footnote{We are particularly grateful to Prof. Kenny Smith from the University of Edinburgh for insightful discussions that inspired this perspective.}.

\section{Experiments}
\label{sec:experiments}

\subsection{Setup}
\label{ssec:experiments:setup}

\textbf{Environment}:
We evaluate tool-use performance using the Berkeley Function Calling Leaderboard V2 (BFCL-V2)~\cite{bfcl2024}, which offers comprehensive coverage of function call types, diverse tasks, programming languages, and executability, and has been widely adopted in recent works~\cite{qwq32b,cohere2025command}.
As our work is the first to investigate LLMs' ability to model environment dynamics, we begin with \emph{single-turn} interactions to ensure a clean and tractable problem formulation, in a serverised BFCL-V2 environment in order to run online RL training. 
Regarding the base model, considering the constraints of our computes, we choose Cohere's R7B, given its leading performance on various agent benchmarks~\cite{cohere2025command}. 

\textbf{SFT Data}:
During the SFT stage, to constitute the function call (fc) SFT dataset $\FcSftDataset$, inspired by~\cite{liu2024apigen, prabhakar2025apigen}, we synthesised pairs of \verb+<prompt,completion>+ following the distribution of BFCL-V2.
Regarding the state prediction (sp) SFT dataset $\SpSftDataset$, we first split the state space $\sZ$ into two subsets:
1) pass states $\sZ^+$ where the completions successfully passed the check of BFCL-V2 and received a score of $1$;
2) error states $\sZ^-$ where the completions failed on the BFCL-V2's check and received a score of $0$.
Given the format of BFCL-V2's return messages, we denote the shared prefix of pass states in $\sZ^+$ as $\passmsg$, and similarly $\errormsg$.
Note that, under this setup, there exhibits a bijection between the BFCL-V2's resulted state subspaces $\{\sZ^+, \sZ^-\}$ and the scores from the reward function $\{0, 1\}$, which we utilise later to truncate the generation when running SVS during inference time.
Following this procedure, we constitute $\SpSftDataset$ of \verb+<prompt,completion,result>+ triplets from our accumulated run-logs of BFCL-V2 tests.

\textbf{RL Data}:
In the following RL stage, to maintain the online RL training and validation distributions as independent and identical, we use $80\%$ from the original BFCL-V2 \emph{prompt} set as the training set, and keep the remaining $20\%$ to validate the generalisation performance.
Note that we intentionally keep at least $20$ test prompts per category in the final validation set, as certain categories contain $\leq50$ samples, thus $20\%$ of them lacks of statistical significance.

\textbf{SVS Scoring Function:}
During the inference time, following GenRM~\cite{zhang2024generative}, we use a scoring function $\textit{score}$ in the following Equation~\ref{eq:experiments:setup:metric_func} to run SVS illustrated in Algorithm~\ref{alg:method:svs}:
\begin{equation}
    \textit{score}\bigl(\policy(\cdot|\vx_1, \hat{\vy}_1), \dots, \policy(\cdot|\vx_k, \hat{\vy}_k)\bigr) \triangleq \arg\max_{j}\left(\left\{\policy(\passmsg|\vx_j, \hat{\vy}_j)\right\}\right).
    \label{eq:experiments:setup:metric_func}
\end{equation}

Examples of all the above types of data are provided in Appendix~\ref{appsec:data}.

\subsection{How proficient is the model at dynamics modelling?}
\label{ssec:experiments:dm_performance}

\begin{table}[t]
    \centering
    \begin{tabular}{c|cc|l}
    \toprule
    \multirow{2}{*}{\textbf{Actual}} & \multicolumn{2}{c|}{\textbf{Predicted}} & \multirow{2}{*}{\textbf{Metrics}} \\
    \cmidrule(lr){2-3}
     & \textbf{Positive} & \textbf{Negative} &  \\
    \midrule
    \textbf{Positive} & 25.40 & 3.56  & Precision: $90.00\% (86.02\% - 93.78\%)$ \\
    \textbf{Negative} & 2.82  & 68.22 & Recall: $87.71\%  (83.46\% - 91.47\%)$ \\
    \midrule
    \multicolumn{1}{c|}{Accuracy} & \multicolumn{2}{c|}{$93.62\% (91.90\% - 95.21\%)$} & F1-score: $88.84\%  (84.72\% - 92.61\%)$ \\
    \bottomrule
    \end{tabular}
    \caption{Confusion matrix of predicting next states by the model $\policymix$ SFTed on both function call dataset \FcSftDataset~and state prediction dataset \SpSftDataset.}
    \label{tab:experiments:dm_performance:confusion_matrix}
    \vspace{-1em}
\end{table}

\begin{figure}[h]
    \centering
    \begin{subcaptionbox}{Precision\label{fig:experiments:dm_performance:metrics_over_RL:precision}}[0.49\textwidth]{
        \centering
        \includegraphics[width=\linewidth]{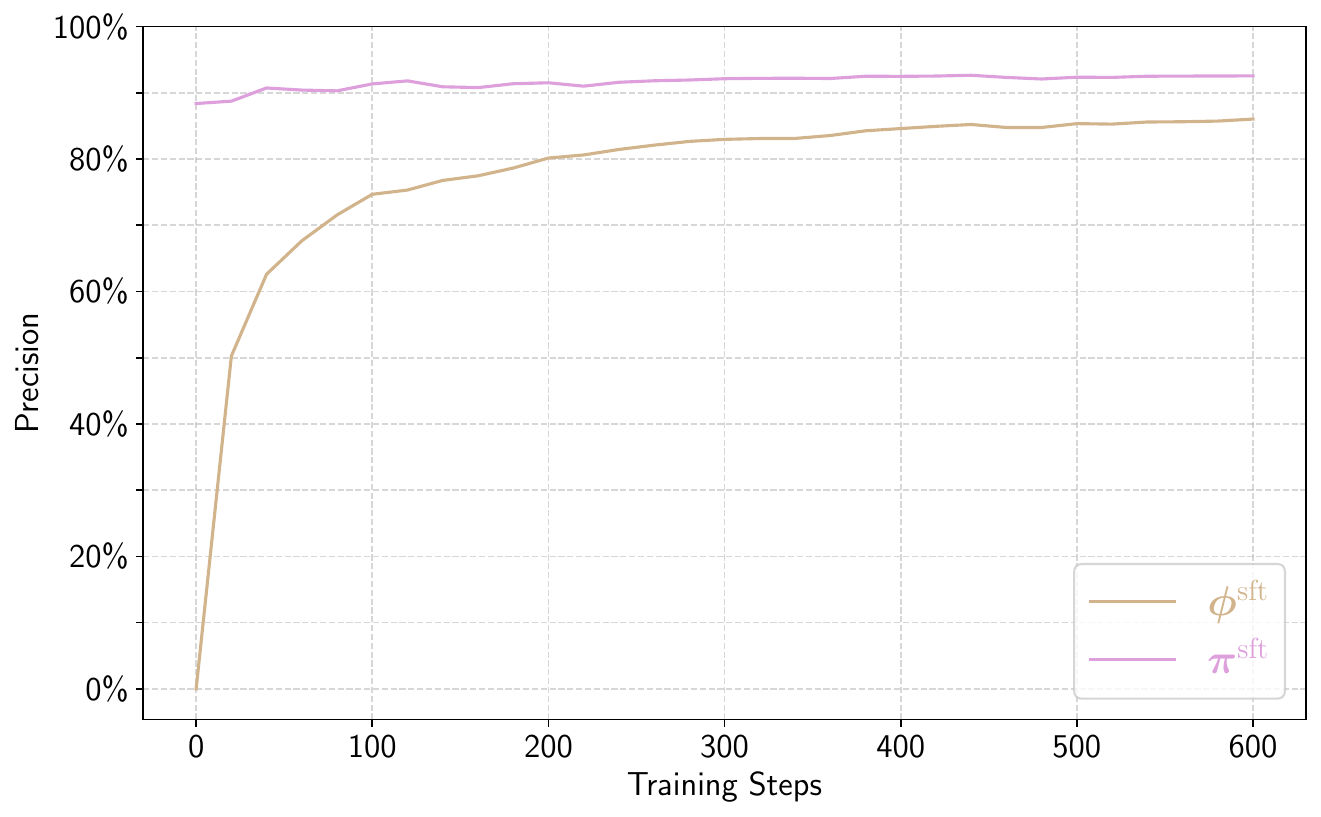}
    }
    \end{subcaptionbox}
    \hfill
    \begin{subcaptionbox}{Recall\label{fig:experiments:dm_performance:metrics_over_RL:recall}}[0.49\textwidth]{
        \centering
        \includegraphics[width=\linewidth]{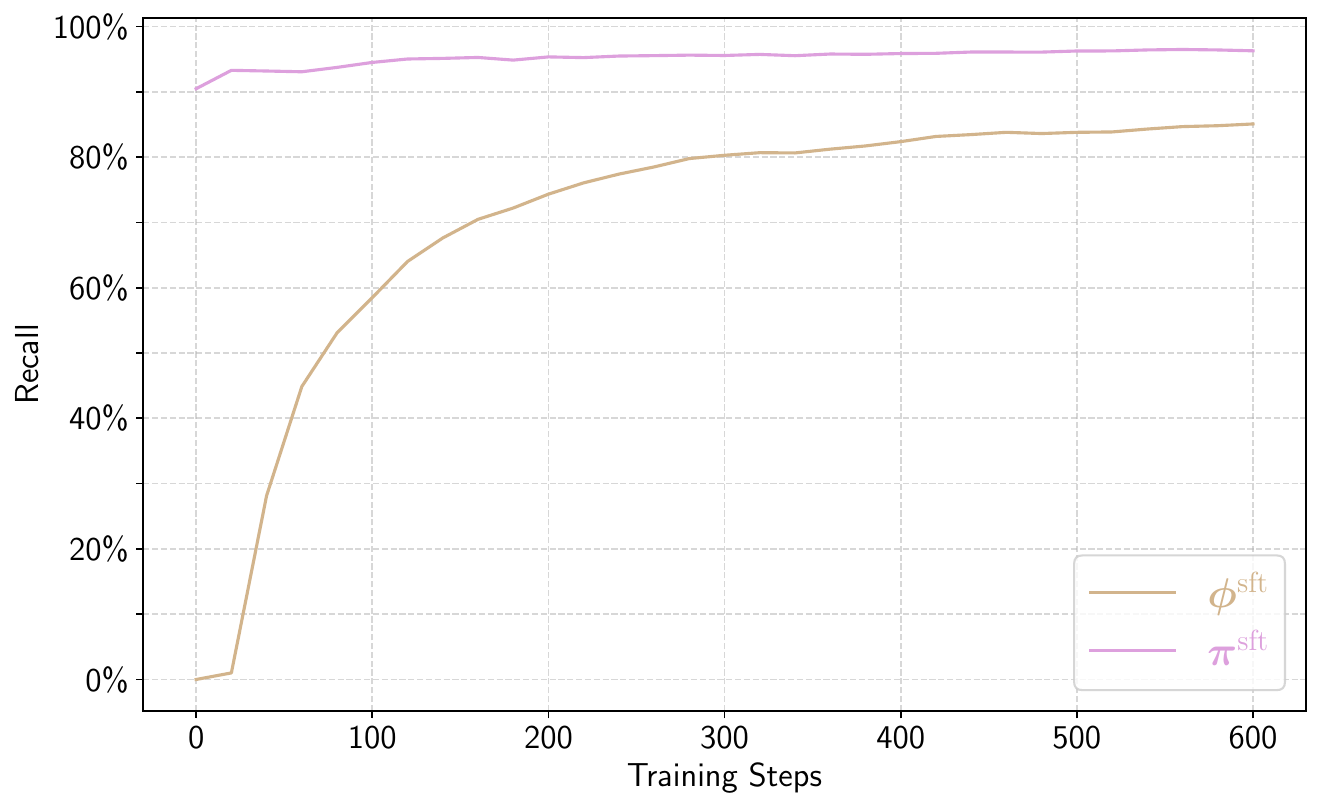}
    }
    \end{subcaptionbox}
    \begin{subcaptionbox}{F1-score\label{fig:experiments:dm_performance:metrics_over_RL:f1}}[0.49\textwidth]{
        \centering
        \includegraphics[width=\linewidth]{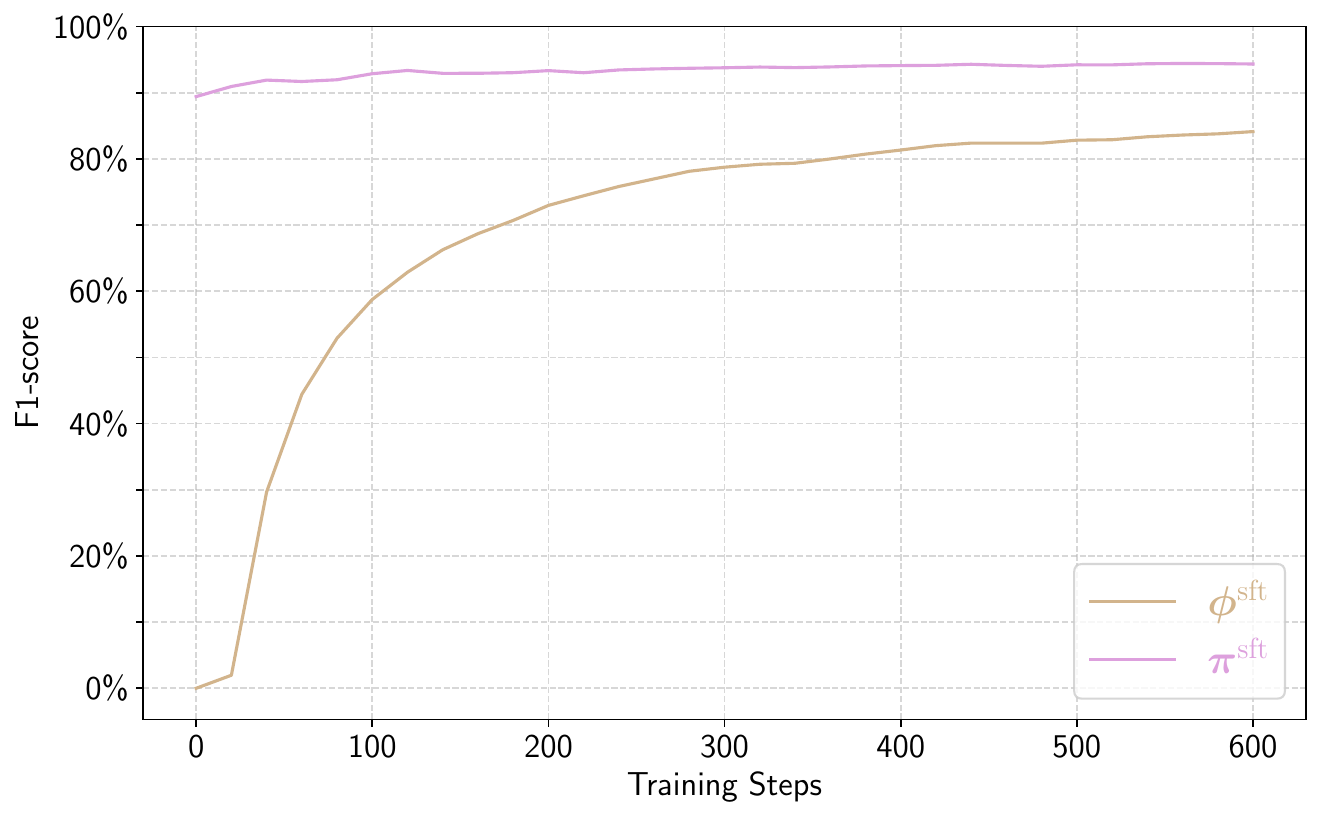}
    }
    \end{subcaptionbox}
    \hfill
    \begin{subcaptionbox}{Accuracy\label{fig:experiments:dm_performance:metrics_over_RL:acc}}[0.49\textwidth]{
        \centering
        \includegraphics[width=\linewidth]{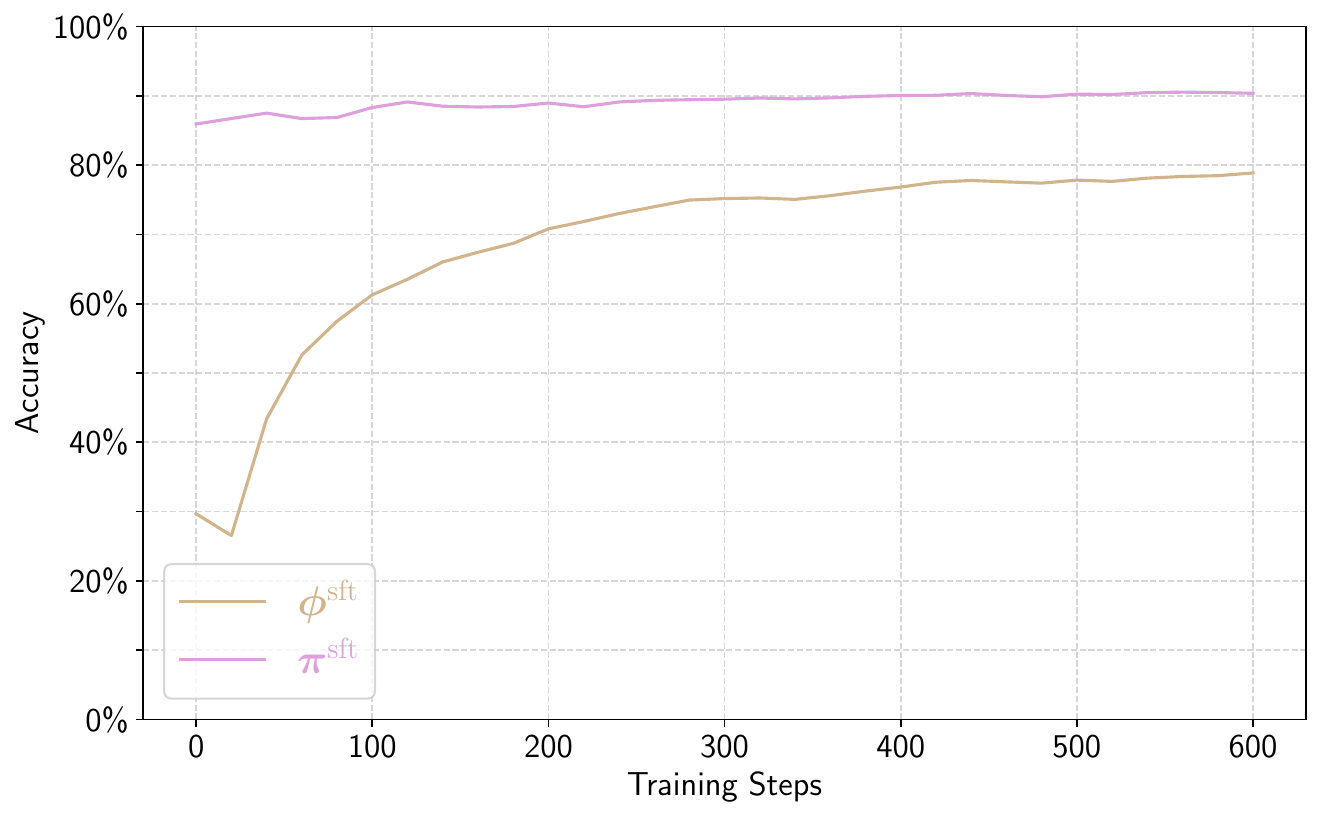}
    }
    \end{subcaptionbox}
    \caption{Performance metrics of results prediction over online RL training with DM loss (Equation~\ref{eq:method:dm:rl:sft_loss}).}
    \label{fig:experiments:dm_performance:metrics_over_RL}
\end{figure}

Since we partition the state space $\sZ$ to $\sZ^+$ and $\sZ^-$, the state prediction task can be framed as a binary classification problem.
In Table~\ref{tab:experiments:dm_performance:confusion_matrix}, we present a detailed breakdown of the model’s performance on the binary classification task formulated in Section~\ref{ssec:experiments:dm_performance}.
The table includes the confusion matrix, with values normalized to sum to $100$ for interpretability, as well as confidence intervals (in brackets) for precision, recall, F1-score, and accuracy.
As can be seen, a model $\policymix$ SFTed on the combined data $\FcSftDataset\cup\SpSftDataset$, achieves a precision of $90.00\%$, recall of $87.71\%$, F1-score of $88.84\%$, and accuracy of $93.62\%$.
Notably, the success rate of this model on BFCL-V2 is only $72.77\%$, which is significantly lower than its discriminative performance, highlighting the gap between accurate state prediction and successful functions calls.
Therefore, a foundation is laid for improving a model's generative capability by leveraging its discriminative capability~\cite{xiong2025self,guo2024oaif}.

We also track these metrics during online RL training with the DM loss function for $\policymix$, and the corresponding curves are shown in Figure~\ref{fig:experiments:dm_performance:metrics_over_RL}.
The red curves represent the model $\policymix$, which is initialized from SFT on $\FcSftDataset\cup\SpSftDataset$, while the blue curves correspond to the baseline model $\policyrag$, fine-tuned only on $\FcSftDataset$.
As shown in the figures, $\policyrag$, which lacks initial state prediction capability, consistently underperforms $\policymix$ across all metrics throughout training.
Even after $600$ steps of RL training, $\policyragrldm$ fails to match the performance of the SFT-only model $\policymix$, which indicates the necessary of the state prediction data.
These results suggest that the benefits of $\SpSftDataset$ cannot be compensated for by relying solely on the downstream \dm~loss during RL training, thus highlighting the necessity of incorporating $\SpSftDataset$ in the SFT stage.


\subsection{How does dynamics modelling benefit SFT and RL for tool-use?}
\label{ssec:experiments:dm_in_sft}

During the experiments in Section~\ref{ssec:experiments:dm_performance}, we observe that incorporating the additional state prediction data $\SpSftDataset$ also leads to a difference in tool use performance.
We compare the the performance of $\policyrag$ - which can do only tool use - with $\policymix$ - which is  capable of both using tools and predicting next states - across all categories of BFCL-V2.
The results in the ``SFT'' section of Table~\ref{tab:experiments:comprehensive:results}
show that $\policymix$ achieves significant improvements on the ``Irrelevance'' category where the models are not expected to generate function calls when the available tools cannot satisfy the user request.
Since the ``Irrelevance'' is specifically designed to evaluate hallucination of models~\cite{bfcl2024}, these results suggest that incorporating the state prediction task helps mitigate hallucination by LLMs~\cite{maynez2020hallucination}.

\begin{table}[!ht]
    \centering
    \resizebox{\textwidth}{!}{
    \begin{tabular}{llccccccc}
        \toprule
        ~~~~~~\textbf{Model} & ~~~~~~~~~~\textbf{Method} & \textbf{Overall \textcolor{black}{\small \it (UW)}} & \textbf{Overall \textcolor{blue}{\small \it (W)}} & & \textbf{Rel.} & \textbf{Irrel.} & \textbf{AST} & \textbf{Exec} \\
        \midrule
        \multicolumn{3}{l}{\textbf{Baselines}} & \multicolumn{1}{c}{\textcolor{blue}{\small \it \# samples}} & \multicolumn{1}{c}{} & \multicolumn{1}{c}{\textcolor{blue}{\small \it (18)}} & \multicolumn{1}{c}{\textcolor{blue}{\small \it (1122)}} & \multicolumn{1}{c}{\textcolor{blue}{\small \it (2501)}} & \multicolumn{1}{c}{} \\
        GPT-4o~\cite{achiam2023gpt4}        & --    & 82.38 & 82.14 & & \textbf{83.33} & 81.31 & 82.51 & -- \\
        Command-A~\cite{cohere2025command}  & --    & 80.57 & 84.14 & & 72.22 & 86.19 & 83.30 & -- \\
        Command-R7B   & --                  & 70.50 & 76.70 & & 55.56 & 81.02 & 74.92 & -- \\
        xLAM-2~\cite{prabhakar2025apigen}   & --    & 72.36 & 71.69 & & 77.78 & 64.34 & 74.95 & -- \\
        ToolACE-2~\cite{liu2025toolace}     & --    & 81.95 & \textbf{85.49} & & 72.22 & \textbf{90.11} & \textbf{83.51} & -- \\
        Watt-tool~\cite{shi2024direct}     & --       & \textbf{82.54} & 81.76 & & \textbf{83.33} & 83.15 & 81.13 & -- \\
        BigAgent~\cite{bitagent8b}      & --       & 82.27 & 81.50 & & \textbf{83.33} & 82.38 & 81.10 & -- \\
        \midrule
        \multicolumn{3}{l}{\textbf{SFT}} & \multicolumn{1}{c}{\textcolor{blue}{\small \it \# samples}} & \multicolumn{1}{c}{} & \multicolumn{1}{c}{\textcolor{blue}{\small \it (18})} & \multicolumn{1}{c}{\textcolor{blue}{\small \it (1122)}} & \multicolumn{1}{c}{\textcolor{blue}{\small \it (2501)}} & \multicolumn{1}{c}{} \\
        \policyrag  & $\FcSftDataset$ only & 66.35    & 66.50 & & 70.73 & 58.05 & 70.26 & 76.25 \\
        \policymix  & $\FcSftDataset\cup\SpSftDataset$  & 70.87    & 73.89 & & 63.41 & 76.32 & 72.88 & 77.53 \\
        \midrule
        \multicolumn{3}{l}{\textbf{SFT + RL}} & \multicolumn{1}{c}{\textcolor{blue}{\small \it \# samples}} & \multicolumn{1}{c}{} & \multicolumn{1}{c}{\textcolor{blue}{\small \it(20)}} & \multicolumn{1}{c}{\textcolor{blue}{\small \it (206)}} & \multicolumn{1}{c}{\textcolor{blue}{\small \it (457)}} & \multicolumn{1}{c}{} \\
        \policyragrl      & \policyrag$\rightarrow$RLOO       & 80.31 & 80.22 & & 75.00 & 89.81 & 76.13 & 96.25 \\
        \policyragrldm    & \policyrag$\rightarrow$RLOO + \dm & 82.13 & 83.16 & & 75.00 & \textbf{91.75} & 79.65 & \textbf{97.50} \\
        \policymixrl      & \policymix$\rightarrow$RLOO       & 81.23 & 81.99 & & 75.00 & 90.00 & 78.68 & 96.25 \\
        \policymixrldm    & \policymix$\rightarrow$RLOO + \dm & \textbf{83.62} & \textbf{86.68} & & 75.00 & 90.29 & \textbf{85.56} & 96.25 \\
        \midrule
        \multicolumn{3}{l}{\textbf{SFT + RL + SVS (with $k$ candidates)}} & \multicolumn{1}{c}{\textcolor{blue}{\small \it \# samples}} & \multicolumn{1}{c}{} & \multicolumn{1}{c}{\textcolor{blue}{\small \it (20)}} & \multicolumn{1}{c}{\textcolor{blue}{\small \it (206)}} & \multicolumn{1}{c}{\textcolor{blue}{\small \it (457)}} & \multicolumn{1}{c}{} \\
        \policymixrldm & \policymixrldm$\rightarrow$ SVS with $k=1$   & 85.77 & 84.26 & & 88.20 & 85.65 & 83.46 & 96.33 \\
        \policymixrldm & \policymixrldm$\rightarrow$ SVS with $k=2$   & 88.20 & 86.71 & & 91.30 & 86.86 & 86.44 & \textbf{96.55} \\
        \policymixrldm & \policymixrldm$\rightarrow$ SVS with $k=4$   & 88.94 & 87.67 & & 91.80 & 87.41 & 87.61 & 96.45 \\
        \policymixrldm & \policymixrldm$\rightarrow$ SVS with $k=8$   & 89.73 & 88.18 & & 93.10 & 88.10 & 88.00 & 96.25 \\
        \policymixrldm & \policymixrldm$\rightarrow$ SVS with $k=16$  & 89.90 & 88.29 & & 93.30 & 88.38 & \textbf{88.03} & 96.15 \\
        \policymixrldm & \policymixrldm$\rightarrow$ SVS with $k=32$  & 90.18 & 88.26 & & 94.10 & 88.59 & 87.86 & 96.25 \\
        \policymixrldm & \policymixrldm$\rightarrow$ SVS with $k=64$  & \textbf{90.69} & \textbf{88.43} & & \textbf{95.00} & \textbf{89.32} & 87.75 & 96.25 \\

        \bottomrule
    \end{tabular}
    }
    \caption{
        Comprehensive category-wise performance comparison across baselines, SFT, SFT+RL, and SFT+RL+SVS models, on BFCL-V2.
        For each section, the number of evaluation examples per column is shown in the second row.
        \textcolor{blue}{\it (W)} indicates metrics weighted by the number of samples, whereas {\it UW} indicates unweighted. Missing results are marked as `--`.
        The ``Exec'' column is provided to show the improvement from RL training on it, but is never counted for the overall performance.
    }
    \label{tab:experiments:comprehensive:results}
    \vspace{-1.5em}
\end{table}

Similarly, we compare the two models — $\policyrag$ and $\policymix$ — both fine-tuned by online RL with and without our \dm~loss, resulting in four variants: $\policymixrldm$, $\policyragrldm$ (with \dm), and $\policymixrl$, $\policyragrl$ (without \dm).
Take $\policymixrldm$ for example, the model is first SFTed on $\FcSftDataset\cup\SpSftDataset$ (thus notated as $\pi$), then further fine-tuned by online RL together with \dm~loss (thus superscripted by ``rd'').
The ``SFT + RL'' section of Table~\ref{tab:experiments:comprehensive:results} shows the success rates of these models across different BFCL-V2 categories.
For analytical clarity, we preserve the ``Exec'' category to show the substantial performance gains over it due to RL training.
The results indicate that incorporating the \dm~loss yields a $>5\%$ improvement in success rate over the AST category, contributing to an overall performance boost.


\subsection{How does the RL/SFT models perform when scaling up test-time compute?}
\label{ssec:experiments:rl_sft_pass_at_k}

\begin{figure}[h]
    \centering
    \includegraphics[width=\textwidth]{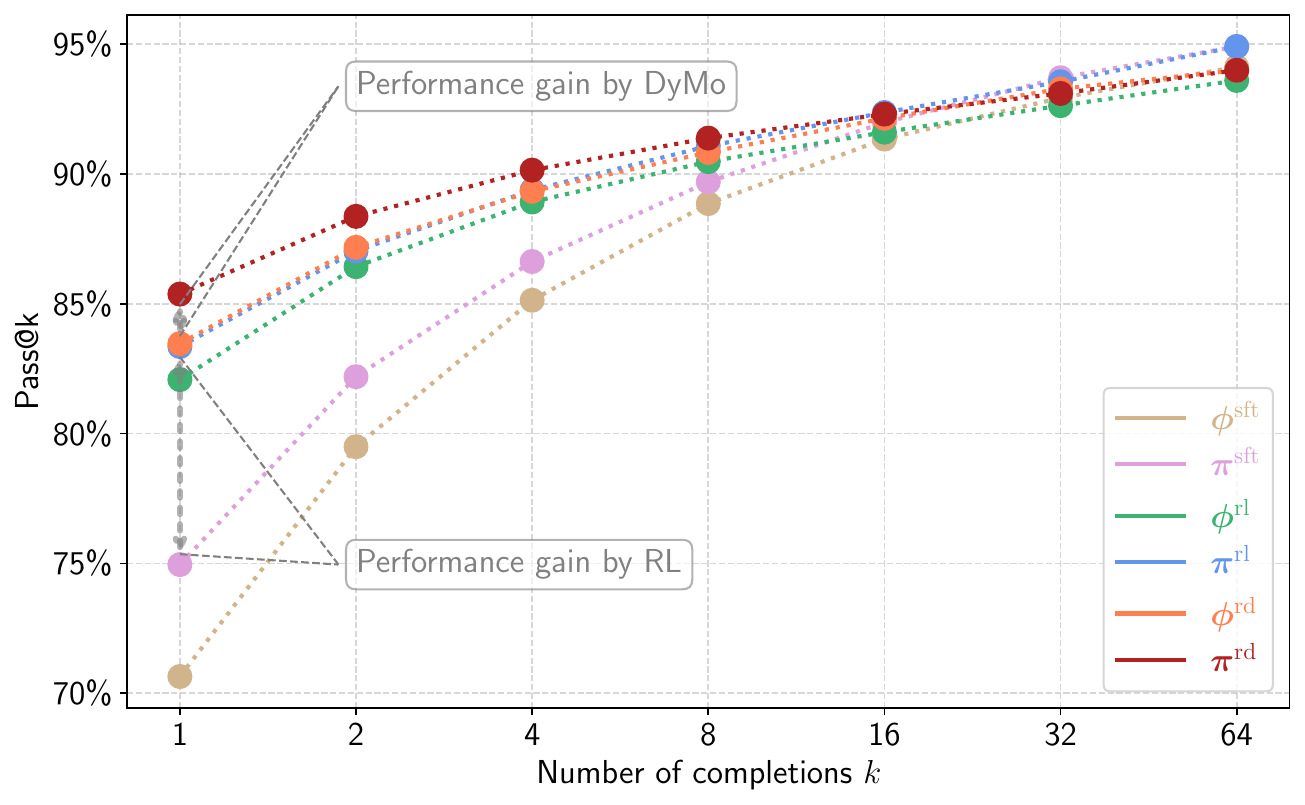}
    \captionof{figure}{Pass@$k$ of all methods investigated in this work scaling over number of completions $k$ during inference time on BFCL-V2. 
    }
    \label{fig:experiments:pass_at_k_all}
\end{figure}

Since the results in the ``SFT'' and ``SFT + RL'' sections of Table~\ref{tab:experiments:comprehensive:results} are based on a greedy decoding strategy, we further examine whether and how the on-policy distribution over completions for a given prompt, i.e., $\policy(\cdot|\vx)$, changes under different training pipelines.
We begin by analysing the impact of online RL training, comparing the RL-trained models — $\policymixrl$ and $\policyragrl$ — with their corresponding SFT-only baselines — $\policymix$ and $\policyrag$ — using the number of completions per request as the variable.
In Figure~\ref{fig:experiments:pass_at_k_all} and Figure~\ref{fig:experiments:pass_hat_k_all}, we report pass@$k$ and pass\textasciicircum$k$~\cite{yao2025taubench} respectively as the evaluation metrics. 

As indicated by the ``performance gain by RL'' in Figure~\ref{fig:experiments:pass_at_k_all}, online RL significantly improves pass@$k$ when $k \leq 8$.
More importantly, as indicated by the ``performance gain by RL'' in Figure~\ref{fig:experiments:pass_hat_k_all}, online RL consistently improves pass\textasciicircum$k$ over all $k$ values, as evidenced by the consistent gap between the RL-trained models — $\policymixrl$ and $\policyragrl$ — over their SFT-only counterparts — $\policymix$ and $\policyrag$.
These results suggest that the on-policy distributions induced by the RL models yields a more consistent and reliable function calling performance than the distributions induced by the SFT models.

We also note that our pass@$k$ curves align with the findings in mathematical reasoning tasks~\cite{yue2025does}, where Yue et al. conclude that ``base models can achieve a comparable or even higher pass@$k$ score compared to their RL counterparts at large $k$ values''.
However, in our setup, we found that base model can hardly match pass@k or pass\textasciicircum$k$ of SFT and RL models, which we argue is due to that correct function calls are sparser to generate.


\subsection{How does dynamics modelling impact the test-time compute scaling of RL models?}
\label{ssec:experiments:dm_for_rl_scale_up}

\begin{figure}[t]
    \centering
    \includegraphics[width=\textwidth]{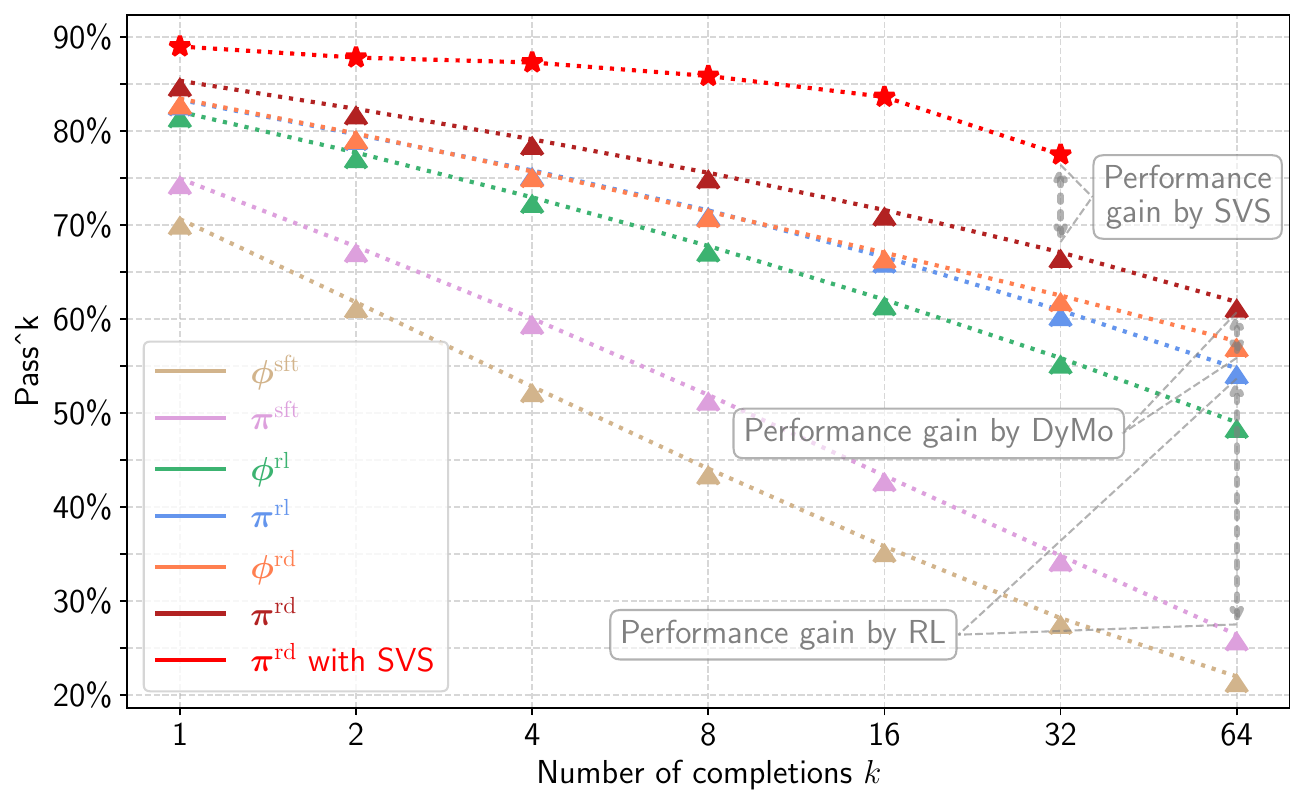}
    \captionof{figure}{Pass\textasciicircum$k$ of all methods investigated in this work scaling over number of completions $k$ during inference time on BFCL-V2.
    More details about the ``performance gain by SVS'' are provided in Section~\ref{ssec:experiments:svs_scale_up} and Table~\ref{tab:experiments:svs_scale_up}.
    }
    \label{fig:experiments:pass_hat_k_all}
\end{figure}

Building on the observation that RL models achieve higher success rates over test-time compute scales, we further investigate the impact of incorporating the \dm~loss during online RL training.
Similarly, in Figure~\ref{fig:experiments:pass_at_k_all} and Figure~\ref{fig:experiments:pass_hat_k_all}, we report pass@$k$ and pass\textasciicircum$k$ for RL models trained with the \dm~loss — $\policymixrldm$ and $\policyragrldm$ — compared to those trained without it — $\policymixrl$ and $\policyragrl$.

As indicated by the ``performance gain by \dm'' in Figure~\ref{fig:experiments:pass_at_k_all}, adding the \dm~loss during online RL improves pass@$k$ when $k\leq8$.
Meanwhile, as indicated by the ``performance gain by \dm'' in Figure~\ref{fig:experiments:pass_hat_k_all}, \dm~loss also consistently improves pass\textasciicircum$k$ over all numbers of completions per prompt $k$.
Note that SVS is not utilised in the experiments so far, thus the improvements are solely due to the \dm~loss.
More notably, incorporating the \dm~in both the SFT and RL stages results in $\policymixrldm$, which achieves the highest pass\textasciicircum$k$ for all values of $k$.
The consistent gap between pass\textasciicircum$k$ curves of $\policymixrldm$/$\policyragrldm$ and $\policymixrl$/$\policyragrl$ also indicate the \dm~loss can help to further improve the consistency and reliability of function calling performance on top of RL.
These results demonstrate the effectiveness and benefits of integrating \dm~into both the SFT and RL phases. 

\subsection{How does self-verification sampling scale over test-time compute?}
\label{ssec:experiments:svs_scale_up}

So far, we have focused primarily on the benefits of incorporating \dm~during model training.
However, as introduced in Section~\ref{ssec:method:svs}, during inference time, self-verification sampling (SVS) actually unifies the policy (as in model-free RL), the environment model (as in model-based RL), and the value function (under our specific state-space split) into a single LLM.
This paradigm enables the model to scale test-time compute by generating more candidate completions per user request \textbf{without} querying the oracle environment function $\gE$.
To evaluate the effectiveness of SVS, we compare \emph{pass\textasciicircum$k$ with SVS} against \emph{pass\textasciicircum$k$ without SVS} of model $\policymixrldm$.
For pass\textasciicircum$k$ with SVS, we sample $c$ candidates for each trial and $k$ trials per prompt, thus $k\times c$ candidate completions in total for each prompt.
Further, per candidate group for each trial, following GenRM~\cite{snell2025scaling}, we adopt the scoring function defined in Equation~\ref{eq:experiments:setup:metric_func} as the metric to select just one output from the $c$ candidates (thus $k$ outputs in the end).

As shown in Table~\ref{tab:experiments:svs_scale_up}, SVS achieves improved pass\textasciicircum$k$ over all $k$ values, demonstrating that self-verification enables effective scaling with additional computes.
More importantly, the consistent improvement of SVS performance with increasing $k$ highlights our method as a novel test-time compute scaling strategy — one that leverages the model’s internal environment approximation to self-verify and select the most reliable candidate completion.
In Section~\ref{ssec:experiments:refuse} and Section~\ref{sec:discussion}, we provide further insights about our current SVS setup.

\begin{table}[!h]
\centering
\resizebox{\textwidth}{!}{%
\begin{tabular}{cccccccc}
\toprule
\multicolumn{2}{c}{$k$}   & 1  & 2  & 4  & 8 & 16 & 32 \\ \hline
\multirow{3}{*}{pass\textasciicircum$k$} & with SVS  & $89.02\%$ & $87.97\%$ & $87.19\%$ & $86.14\%$ & $84.05\%$ & $78.05\%$ \\
 & ($c$ for each trial) & (64) & (32) & (16) & (8) & (4)  & (2)  \\ \cline{2-8} 
                        & without SVS & $87.68\%$ & $82.38\%$ & $79.14\%$ & $75.58\%$ & $71.61\%$ & $67.11\%$ \\
\bottomrule
\end{tabular}
}
\caption{
Pass\textasciicircum$k$ with and without SVS over $k$ trials in the oracle environment.
Augmented with SVS, per prompt, we first generate $c$ candidate completions for each trial, then select just one to output by the scoring function defined in Equation~\ref{eq:experiments:setup:metric_func} for all $k$ trials.
Therefore, there are $k \times c$ candidates in total for each prompt, by querying the oracle environment also $k$ times as to pass\textasciicircum$k$ without SVS.
}
\label{tab:experiments:svs_scale_up}
\end{table}

Beyond the above experiment, we also compare pass@$k$ of the ``Best-of-$N$'' test-time compute scaling strategy with the \emph{pass@1} performance of $\policymixrldm$ using SVS with $k$ candidate completions per prompt, thus both methods operate under the same inference compute budget.
As shown in the results provided in the ``SFT + RL + SVS'' section of Table~\ref{tab:experiments:comprehensive:results}, increasing number of candidates $k$ in SVS consistently improves the \emph{pass@1}, which demonstrated the effectiveness of the model's internal environment model.
It is unsurprising to observe that querying the model’s internal environment model is less efficient than accessing the oracle environment function under the same compute budget, and should be seen as an upper-bond.
However, we also argue that relying on the oracle may be impractical in many real-world applications involving stateful environments.
For example, the model is not expected to place $k$ parallel orders for a single shopping request or to book $k$ tickets on the same flights for a travel planning request.


\subsection{What if the model is allowed to refuse?}
\label{ssec:experiments:refuse}

\begin{quote}
``I'm sorry, Dave. I'm afraid I can't do that.'' -- 2001: A Space Odyssey
\end{quote}

As may already be observed, a notable limitation of the scoring function defined in Equation~\ref{eq:experiments:setup:metric_func} is that the model is still required to output a completion, even in cases where all candidate completions are self-verified as failed trials.
That is, our model might roll-out $\errormsg$ for all generated candidates.
In such cases, we argue that it is both reasonable and desirable for the model to “refuse” the request by returning a message that informs the user the query cannot be completed reliably. 
Formally, we define the revised scoring function as follow:
\begin{equation}
    \begin{aligned}
        & \textit{score}\bigl(\policymixrldm(\hat{\vz}_1|\vx_1, \hat{\vy}_1), \dots, \policymixrldm(\hat{\vz}_k|\vx_k, \hat{\vy}_k)\bigr)
        \triangleq
        \arg\max_{j}\left(\left\{\policymixrldm(\hat{\vz}_j|\hat{\vy}_j, \vx_j)\right\}\right) \\
        \text{s.t.}~ & ~\passmsg \prec \hat{\vz}_j \text{ and } \policymixrldm(\passmsg|\hat{\vy}_j, \vx_j) > \tau
    \end{aligned}
    \label{eq:experiments:refuse}
\end{equation}
where $\passmsg \prec \hat{\vz}_j$ means that $\hat{\vz}_j$ starts with $\passmsg$ as the prefix, and $\tau$ is the threshold hyperparameter.

\begin{figure}[t]
    \centering
    \includegraphics[width=\textwidth]{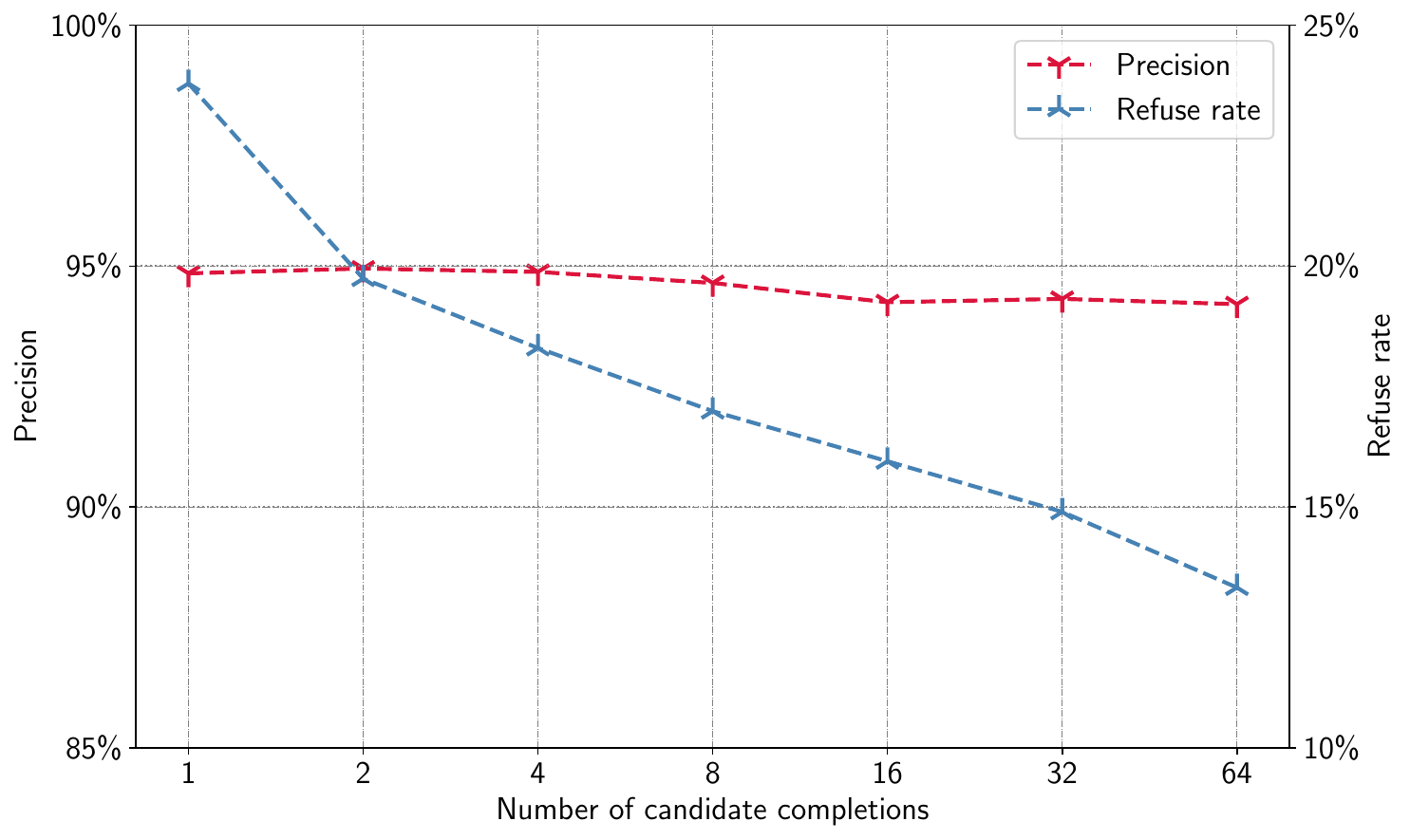}
    \caption{Precision and refuse rate over $k$ candidate completions for self-verification sampling.}
    \label{fig:experiments:refuse:precision_refuse_over_k}
\end{figure}

Using Equation~\ref{eq:experiments:refuse} as the scoring function, the model $\policymixrldm$ classifies a completion $\hat{\vy}$ as positive if and only if $\policymixrldm(\passmsg|\hat{\vy}, \vx)>\tau$; otherwise, it is classified as negative.
By sweeping across a range of thresholds  $\tau\in[0.5,0.99]$, we find that $\tau=0.92$ offers a favourable trade-off between precision and refusal.
Fixing $\tau=0.92$, we then examine how precision and refuse rate vary with the number of candidate completions $k$, as shown in Figure~\ref{fig:experiments:refuse:precision_refuse_over_k}.

Surprisingly, the model maintains a precision of $\sim94.5\%$ across values of $k$, while the refuse rate decreases notably from $23.79\% (k=1)$ to $13.33\% (k=64)$.
Since precision reflects the proportion of correct completions among all non-refused outputs, we interpret it as a proxy for the reliability of the model’s responses.
Under this view, our results suggest that reliability remains stable as the number of candidates increases, while the refusal rate drops significantly — indicating improved solution coverage without sacrificing correctness.
These findings highlight the practical value of combining \dm~with SVS: by generating more candidates, the model achieves higher success rates while maintaining high reliability.
We further discuss the broader implications of this observation in Section~\ref{sec:discussion}.




\section{Discussion}
\label{sec:discussion}

\textbf{Internal environment model by \dm}:
Recent advances in RL have shown that incorporating world models can substantially improve performance in complex domains such as board games~\cite{schultz2024mastering} and video games~\cite{hafer2025dreamer}.
Building on this line of work, our approach takes a further step by unifying the world model and the policy into a single LLM through \dm, and demonstrates the practical benefits of this unification in tool use scenarios.
We also note that similar motivations have emerged in reward modelling, where LLMs are fine-tuned either into stand-alone reward models~\cite{zhang2024generative}, or into generative models capable of self-rewarding by reasoning over multiple steps in a single completion~\cite{xiong2025self}.
Our work extends this broader trend by showing how \dm~can enhance function calling beyond reasoning, particularly in stateful environments.

\textbf{Low true negative ratio problem of SVS}:
In our analysis of the results in Section~\ref{ssec:experiments:svs_scale_up}, we observe that the model $\policymixrldm$ exhibits surprisingly low true negative ratio ($<50\%$ TNR), despite achieving strong precision and recall.
As the model’s success rate increases through online RL training, the proportion of correct completions steadily rises. 
This leads to a highly imbalanced distribution between completions beginning with $\passmsg$ and $\errormsg$,  thus introduces a bias toward predicting states in  $\sZ^+$.
Consequently, we observe that the model tends to ``over-refuse'' its own completions, i.e. it incorrectly verifies many correct completions as failures via its internal environment model.
A straightforward mitigation strategy would be to incorporate additional negative samples from $\FcSftDataset$, thereby exposing the model to a more balanced distribution during \dm~in RL training.
Due to time and research scope constraints, we leave this direction for future work.
Nonetheless, we argue that our method significantly enhances the reliability of model outputs: higher precision implies that completions self-verified by the model are more likely to be correct.
This property is particularly valuable in high-stakes or safety-critical domains, such as healthcare or finance, where even a few incorrect outputs can lead to undesirable or irreversible outcomes.

\textbf{Test-time compute scaling via \dm~and SVS}:
As discussed in previous sections, the proposed \dm~and SVS provide a novel strategy for test-time compute scaling.
Here, we offer additional reflections from both data-centric and modelling perspectives.
First, we highlight that \dm~can benefit from failed completions, since a complete environment model should be capable of handling both successful and failed trajectories.
Given the vast amount of run logs accumulated from software systems over decades, we argue that \dm~unlocks a largely under-explored data source: rich, naturally occurring software run logs.
In particular, the ability of \dm~to learn from failed completions helps improve the fidelity of the internal environment model — a capability, to the best of our knowledge, not explicitly addressed in prior works from the LLM community.
Secondly, from the perspective of world modelling, we hypothesise that programs are often written with an implicit and internal world model of the developers who coded upon assumptions about environment dynamics, constraints, and expected behaviours.
These implicit world models are then reflected in the run logs, which can then be captured and fitted by the proposed \dm~method.
Through SVS, this learned environment model can be exploited at test time, enabling the model to improve its decision quality without external feedback.
While this hypothesis is promising, a deeper exploration of the relationship between program execution and world modelling lies beyond the scope of this work, and we leave it for future investigation.

\section{Conclusion}
\label{sec:conclusion}

In this work, we investigate the challenge of tool use in stateful environments, where existing test-time compute strategies become impractical due to repeated environment queries.
To address this, we propose \dm, a method that augments LLM fine-tuning with an additional state prediction task during both the SFT and RL stages, enabling a next-state prediction capability of the model.
Experiments on the BFCL-V2 benchmark show that incorporating \dm~significantly reduces hallucinations during SFT and improves the success rate over RL training loops.
Notably, we also observe that RL models consistently outperform SFT models in mitigating hallucinations.
Furthermore, we demonstrate that correct tool calls are retrievable for over $93\%$ of prompts using a parallel Best-of-$N$ decoding strategy, indicating that both SFT and RL models have learned sufficiently expressive on-policy distributions.
Building on this insight, we introduce a self-verification sampling (SVS) strategy, which consistently improves pass\textasciicircum$k$ and pass@1 performance by leveraging the model’s internal environment model.
Crucially, by allowing the model to refuse uncertain completions, our approach produces more reliable outputs in scenarios where correctness is essential.
Overall, our findings highlight a promising direction for extending planning algorithms from the RL community to LLMs in dynamic and stateful environments.


\begin{ack}
We would like to express our sincere gratitude to Kenny Smith and Yao Zhao for their insightful discussions on refining the initial design of our method.
We are also gratefully thankful to Minjie Xu, Yixuan Su, and Patrick Lewis for their valuable assistance with the data used in our experiments.
We are particularly indebted to Matthieu Geist for his constructive feedback and thoughtful suggestions, which substantially contributed to improving the quality of this work.

We further acknowledge the significant efforts of the Reinforcement Learning team at Cohere, especially Yannis Flet-Berliac, Nathan Grinsztajn, Pierre Clavier, Eugene Choi, Yash Chandak, and Mohammad Gheshlaghi Azar,for their dedication in developing the infrastructure necessary for this project.
In particular, we extend our deep appreciation to Eugene Tarassov and Elena Tommasone for their extensive support in implementing the systems required to run our algorithms.
\end{ack}


\newpage
\bibliography{citations}
\bibliographystyle{unsrtnat}


\newpage
\appendix

\section{Examples of Data Used for Training LLMs}
\label{appsec:data}

In this section, we present examples of the datasets curated for supervised fine-tuning (SFT) of large language models (LLMs) on the tool use and state prediction tasks, as described in Section~\ref{sssec:method:dm:sft} and Section~\ref{ssec:experiments:setup}.

\subsection{Example of Function Call Supervised Fine-tuning Dataset $\FcSftDataset$}
\label{appssec:data:fc_sft}

Below, we provide an example of the function call SFT data.
The completion shown in the yellow box corresponds to the ground-truth output used for supervised training and is guaranteed to be correct.
Importantly, our function call SFT dataset $\FcSftDataset$ does not include any data from the original BFCL benchmark; the following example is provided solely for illustrative purposes.

\begin{tcolorbox}[colback=gray!10, breakable, colframe=gray!50, coltitle=black, boxrule=0.5pt, arc=1pt, title=\textbf{Example 1}: Humidity Forecast Query]

    \begin{preamblebox}{System Preamble}
        You are a large language model AI assistant. Your knowledge cutoff date is ...
       
        ...
        
        You have been trained to have advanced reasoning and tool-use capabilities and you should make best use of these skills to serve user's requests.

        Here is the list of tools that you have available to you.
        You can ONLY use the tools listed here. When a tool is not listed below, it is NOT available and you should NEVER attempt to use it.
        Each tool is represented as a JSON object with fields like "name", "description", "parameters" (per JSON Schema), and optionally, "responses" (per JSON Schema).

        \begin{minipage}{\linewidth}
            {\footnotesize\texttt{[\\
                \{"name": "weather.humidity\_forecast", "description": "Retrieve a humidity forecast for a specific location and time frame.", "parameters": \{"type": "object", "properties": \{"location": \{"type": "string", "description": "The city that you want to get the humidity for."\}, "days": \{"type": "integer", "description": "Number of days for the forecast."\}, "min\_humidity": \{"type": "integer", "description": "Minimum level of humidity (in percentage) to filter the result. Optional parameter. Default is 0."\}\}, "required": ["location", "days"]\}, "responses": null\},\\
                \{"name": "get\_team\_score", "description": "Retrieves the latest game score, individual player stats, and team stats for a specified sports team.", "parameters": \{"type": "object", "properties": \{"team\_name": \{"type": "string", "description": "The name of the sports team."\}, "league": \{"type": "string", "description": "The league that the team is part of."\}, "include\_player\_stats": \{"type": "boolean", "default": false, "description": "Indicates if individual player statistics should be included in the result. Default is false."\}\}, "required": ["team\_name", "league"]\}, "responses": null\}\\
            ]}}
        \end{minipage}

        ...

    \end{preamblebox}

    \begin{promptbox}{User Prompt}
        What is the humidity level in Miami, Florida in the upcoming 7 days?
    \end{promptbox}

    \begin{completionbox}{Completion}
        I'll query the \verb+weather.humidity_forecast+ to give user an answer. The call is

        \begin{minipage}{\linewidth}
            {\footnotesize
            \texttt{
                \{"weather.humidity\_forecast": \{"location": ["Miami", "Miami, Florida", "FL"], "days": [7], "min\_humidity": ["", 0]\}\}
            }}
        \end{minipage}
        
    \end{completionbox}

\end{tcolorbox}

\subsection{Examples of State Prediction Supervised Fine-tuning Dataset $\SpSftDataset$}
\label{appssec:data:sp_sft}

Below, we present an example of the state prediction SFT data.
Example 2 corresponds to a previously successful call to the humidity forecast function.

\begin{tcolorbox}[colback=gray!10, breakable, colframe=gray!50, coltitle=black, boxrule=0.5pt, arc=1pt, title=\textbf{Example 2}: A Successful Call to the Humidity Forecast Query]

    \begin{preamblebox}{System Preamble}
        You are a large language model AI assistant. Your knowledge cutoff date is ...
        
        (Identitcal to exmpale 1 above, thus we skip the following content to save pages.)
    \end{preamblebox}

    \begin{promptbox}{User Prompt}
        What is the humidity level in Miami, Florida in the upcoming 7 days?
    \end{promptbox}

    \begin{completionbox}{Completion}
        I'll query the \verb+weather.humidity_forecast+ to give user an answer. The call is

        \begin{minipage}{\linewidth}
            {\footnotesize
            \texttt{
                \{"weather.humidity\_forecast": \{"location": ["Miami", "Miami, Florida", "FL"], "days": [7], "min\_humidity": ["", 0]\}\}
            }}
        \end{minipage}
        
    \end{completionbox}

    \begin{passbox}{Pass}
        \textcolor{green}{<|pass|>}
        
        \begin{minipage}{\linewidth}
            {\footnotesize
            \texttt{
                \{ 
                    "status": 1,
                    "forecast": [
                        \{ "date": "2025-04-11", "min\_humidity": 62, "max\_humidity": 78 \},
                        ...
                    ]
                \}
            }}
        \end{minipage}
    \end{passbox}

\end{tcolorbox}

Example 3 corresponds to a previously failed call to the humidity forecast function and is used as a negative example in the state prediction task.

\begin{tcolorbox}[colback=gray!10, breakable, colframe=gray!50, coltitle=black, boxrule=0.5pt, arc=1pt, title=\textbf{Example 3}: A Failed Call to the Humidity Forecast Query]

    \begin{preamblebox}{System Preamble}
        You are a large language model AI assistant. Your knowledge cutoff date is ...
        
        (Identitcal to exmpale 1 above, thus we skip the following content to save pages.)
    \end{preamblebox}

    \begin{promptbox}{User Prompt}
        What is the humidity level in Miami, Florida in the upcoming 7 days?
    \end{promptbox}

    \begin{completionbox}{Completion}
        I'll query the \verb+weather.humidity_forecast+ to give user an answer. The call is

        \begin{minipage}{\linewidth}
            {\footnotesize
            \texttt{
                \{"weather.humidity\_forecast": \{"location": ["Miami", "Miami, Florida", "FL"], "days": [-7], "min\_humidity": ["", 0]\}\}
            }}
        \end{minipage}
        
    \end{completionbox}

    \begin{errorbox}{Error}
        \textcolor{red}{<|error|>}
        
        \begin{minipage}{\linewidth}
            {\footnotesize
            \texttt{
                \{ 
                    "status": 0,
                    "error": "Invalid input to argument 'days', expected a positive integer, but got -7...."
                \}
            }}
        \end{minipage}
    \end{errorbox}

\end{tcolorbox}

\end{document}